# Application of Artificial Neural Networks for Investigation of Pressure Filtration Performance, a Zinc Leaching Filter Cake Moisture Modeling

Masoume Kazemi[1], Davood Moradkhani[1✉], Alireza A. Alipour[2]

**Abstract**

Machine Learning (ML) is a powerful tool for material science applications. Artificial Neural Network (ANN) is a machine learning technique that can provide high prediction accuracy. This study aimed to develop an ANN model to predict the cake moisture of the pressure filtration process of zinc production. The cake moisture was influenced by seven parameters: temperature (35 ℃ and 65 ℃), solid concentration (0.2 and 0.38 g/L), pH (2, 3.5, and 5), air-blow time (2, 10, and 15 min), cake thickness (14, 20, 26, and 34 mm), pressure, and filtration time. The study conducted 288 tests using two types of fabrics: polypropylene (S1) and polyester (S2). The ANN model was evaluated by the Coefficient of determination ($R^2$), the Mean Square Error (MSE), and the Mean Absolute Error (MAE) metrics for both datasets. The results showed $R^2$ values of 0.88 and 0.83, MSE values of $6.243 \times 10^{-07}$ and $1.086 \times 10^{-06}$, and MAE values of 0.00056 and 0.00088 for S1 and S2, respectively. These results indicated that the ANN model could predict the cake moisture of pressure filtration in the zinc leaching process with high accuracy.

## 1. Introduction

Zinc hydrometallurgy is an economical and environmentally suitable technique to treat zinc in industrial processes. It consists of three main sub-processes: leaching, purification, and electro-wining [1]. In the purification sub-process, zinc powder is used to remove impurities from the leach solution. These impurities form a solid residue, called filter cake, which contains iron, nickel, cobalt, and zinc and can be separated from the leach solution.

---

1 Zanjan University, Department of Materials Science and Engineering
2 Institute for Advanced Studies in Basic Sciences, Department of Computer Science
* Corresponding author: Faculty of Engineering Zanjan University, Zanjan, Iran
E-mail address: moradkhani@znu.ac.ir Corresponding author: Faculty of Engineering Zanjan University, Zanjan, Iran

The presence of zinc in the filter cake causes significant financial losses, as it reduces the yield of zinc production.

The amount of zinc in the filter cake in the leaching process may depend on several factors, such as 1) the type and composition of the filter cake. Different filter cakes may have different zinc-containing phases and leaching behaviors. 2) The storage conditions of the filter cake. The filter cake may undergo oxidation or hydration during storage, which may affect its leaching performance. 3) The leaching parameters, such as acid concentration, acid-to-solid ratio, leaching time, temperature, etc. These parameters may influence the dissolution and extraction of zinc from the filter cake [2]. Therefore, it is important to optimize the conditions of the leach solution and minimize the zinc losses in the purification sub-process. This would require a reliable control scheme to ensure the stability and efficiency of the zinc hydrometallurgy process.

With the advent of the fourth industrial revolution, Artificial Intelligence (AI) and Machine Learning (ML) are increasingly used to accelerate research and development in material science. One of the most reliable computational prediction methods that has been widely applied for solving problems and predicting outcomes in the hydrometallurgy industry is Artificial Neural Network (ANN). For example, Wu et al [3] used ANN to optimize the electrolytic process in zinc hydrometallurgy and obtained high-purity metallic zinc with significant economic benefits. Similarly, ANN was applied to predict the effects of operational parameters on the dissolution of Cu, Mo, and Re from molybdenite concentrate by meso acidophilic bioleaching [4]. Liu et al [5] reviewed the state-of-the-art literature on ANN models for the constitutive modeling of composite materials, focusing on discovering unknown constitutive laws and accelerating multiscale modeling. Ebrahimzade et al [6] investigated the kinetic study of valuable metals recovery from waste Lithium-Ion Batteries (LIBs) using ANN. Qing-cui et al [7] developed an ANN model to predict the oxidation of Refractory Gold Concentrate (RGC) by ozone and ferric ions. In our previous study, we successfully developed and evaluated support vector regression (SVR) and random forest regression (RFR) models for predicting the cake moisture content of a pressure filtration process [8].

In this work, we studied the modeling of pressure filtration using ANN with different parameters, such as solid concentration, temperature, pH, air-blow time, pressure, filtration time, and cake thickness. We used these parameters to model the cake moisture and evaluate the relative importance of each variable. This can help to quickly adjust the process to the desired conditions.

## 1.1. Artificial Neural Networks

Artificial Neural Networks (ANNs) are a branch of machine learning models that mimic the human nervous system. An ANN is composed of a collection of connected units or nodes called artificial neurons, which can receive, process, and transmit signals to other neurons. Each connection has a weight that determines the strength of the signal, and each neuron has a threshold that determines whether it activates or not. Typically, neurons are organized into layers, such as an input layer, one or more hidden layers, and an output layer. The input layer receives the raw data, and the output layer produces the desired result. The hidden layers can perform various transformations on the data using the non-linear activation function [9]. ANNs can be classified into supervised and unsupervised learning based on their learning process [10]. There are many types of ANNs, depending on their structure, function, and learning algorithm. In this research, the Multilayer Perceptron (MLP) was selected.

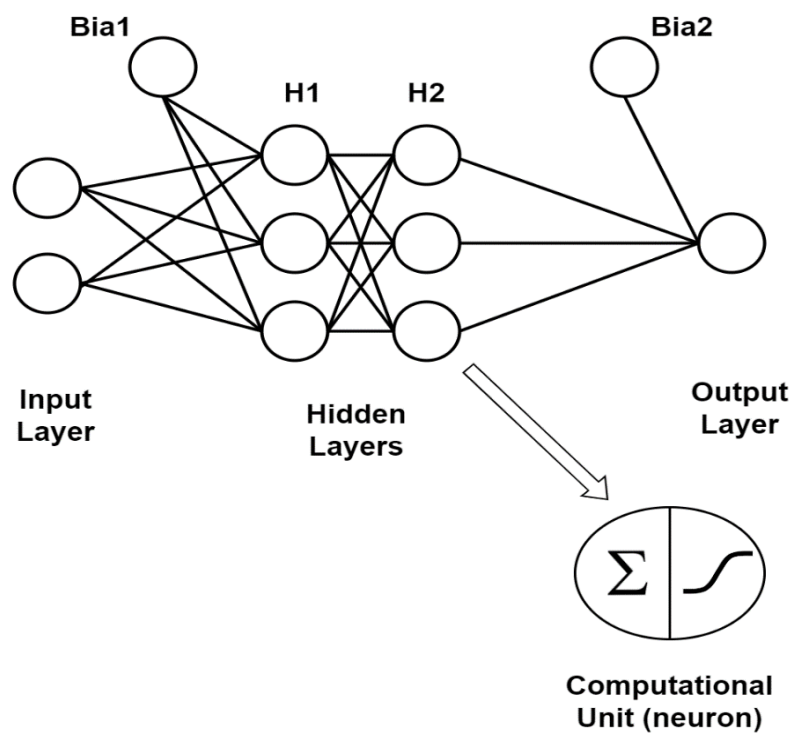

Figure 1 Three-layered feed-forward neural network with one input layer, one (or more) hidden layer(s), and one output layer.

### 1.1.1 Multilayer Perceptron

A Multilayer Perceptron (MLP) is a type of artificial neural network that consists of multiple layers of nodes, where each node is a perceptron, which is a binary classifier that uses a linear function and a threshold to decide whether to activate or not. An MLP has at least three layers: an input layer, a hidden layer, and an output layer. In Figure 1, the input layer receives the raw data, and the output layer produces the desired result. The hidden layers in between can perform various transformations on the data. Unlike a single-layer perceptron, an MLP can use non-linear activation functions at each layer, such as sigmoid, tanh, or ReLU, to learn more complex and non-linear patterns [11]. An MLP can be trained using backpropagation, which is an algorithm that adjusts the weights of the connections based on the error between the predicted output and the actual output. An MLP can be used for both regression and classification tasks, such as image recognition, natural language processing, etc.

The rest of this paper is structured as follows: Section 2 provides an overview of the experiments and methods used in this study. Section 3 presents the results of the ANN model that was developed for pressure filtration of the leaching process. Section 4 concludes the paper with some remarks.

## 2. Material and methods

### 2.1. Experiment

We used zinc concentrates from the Calcimine company in Zanjan province, Iran, with 25% primary moisture, to perform 288 experiments on pressure filtration. These experiments were divided into two groups of 144 one, using polypropylene (S1) and polyester (S2) fabrics.

Table 1 Ranges for input parameters for the filtration process

| Parameter | Level | | | |
|---|---|---|---|---|
| | 1 | 2 | 3 | 4 |
| Solid concentration (g/L) | 0.2 | 0.38 | - | - |
| Temperature (°C) | 35 | 65 | - | - |
| pH | 2 | 3.5 | 5 | - |
| Air-blow time (min) | 2 | 10 | 15 | - |
| Cake thickness (mm) | 14 | 20 | 26 | 34 |

We also varied six parameters in each experiment: solid concentration, temperature, pH, air-blow time, cake thickness, and filtration time. To prepare the zinc concentrate for leaching, a ball mill was used to crush it into powder. Then, the 37.5 kg of powder was leached with sulfuric acid in a 200L reactor, using 125L and 62.5L water to achieve solid: liquid ratios of 0.2 or 0.38 g/L, respectively. The temperature (35 °C or 65 °C) and the pH (2, 3.5, or 5) of the solution were adjusted according to the experimental design. The reactor had a mechanical stirrer with a controller unit and was aerated from the bottom. We used filter press plates of different sizes to obtain cake thicknesses of 14, 20, 26, and 34 millimeters. The filtration time was based on the leaching conditions. In the next stage, the solution was separated from the solid material by filtration. After filtration, the filter cake was dried in an oven at 110-120 °C. Then, the actual cake moisture was calculated. The ranges of these parameters are shown in Table 1.

## 2.2. Modeling

The dataset used for developing an Artificial Neural Networks (ANNs) model in this study was obtained from pressure filtration of the leaching process in Zanjan University's pilot. The dataset has 288 data points, with 144 using polypropylene fabric and 144 using polyester. The input variables are temperature, pH, solid concentration, pressure, air-blow time, filtration time, and cake thickness. The output parameter is the cake moisture of the leaching process.

A set of 100 data (70% of the total) as training data and the remaining 44 data (30%) for testing the ANN model were selected. We set the experimental conditions and results as the input matrix and the target matrix, respectively. The preprocessing of input data was carried out by normalizing in order to improve the performance of the neural network. All input data were normalized according to Eq. (1):

$$N_p = (A_p - A_{mean, p}) / A_{std, p} \qquad (1)$$

where $A_p$ refers to the actual parameter; $A_{mean, p}$ is the mean of actual parameters; $A_{std, p}$ is the standard deviation of actual parameter and $N_p$ is the normalized parameter (model's inputs). The performance of the developed ANN model was evaluated by three parameters: The Coefficient of determination ($R^2$), the Mean Square Error (MSE), and the Mean Absolute Error (MAE). These metrics show how well the Multilayer Perceptron (MLP) matches the modeled output and the measured datasets.

## 3. Results

Figure 2 indicates a scatter plot of the experimental and predicted values of the Artificial Neural Networks (ANNs). The purpose of these graphs is to compare the ANN outputs with the actual values and measure the error between them. Figure 2 indicates that the data points from the ANN model are close to the equality line (y=x), indicating a good fit between the predicted and target values. The further the points deviate from this line, the higher the error of the model.

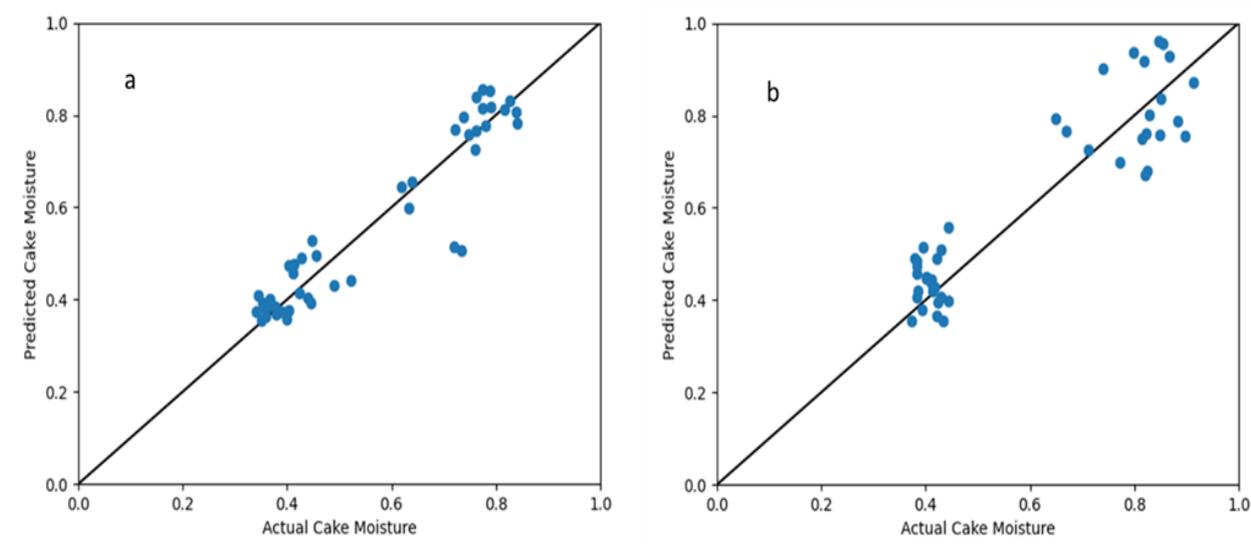

Figure 2. Comparison of actual cake moisture with the predicted cake moisture for the testing dataset: (a) S1; (b) S2.

To evaluate the accuracy of the ANN model, three performance metrics were calculated: The Coefficient of Determination ($R^2$), the Mean Square Error (MSE), and the Mean Absolute Error (MAE). Table 2 shows that the $R^2$ values for S1 and S2 datasets were 0.88 and 0.83, respectively. This indicates that the ANN model had a high prediction accuracy, as $R^2$ values close to 1 imply a good fit between the predicted and actual values. Table 2 shows the error rates between the ANN outputs and the actual values. The MSE values for S1 and S2 were $6.243 \times 10^{-07}$ and $1.086 \times 10^{-06}$, respectively, and the MAE values for S1 and S2 were 0.00056 and 0.00088, respectively. These low error rates suggest that the ANN model was able to predict the cake moisture of the leaching process with high precision. Moreover, the closeness of the data points to the equality line in Figure 2 confirms the low error rates of the ANN model.

Table 2: Summary of ANN model performance

| Dataset | $R^2$ | MSE | MAE |
|---|---|---|---|
| S1 | 0.880 | $6.243 \times 10^{-07}$ | 0.00056 |
| S2 | 0.838 | $1.086 \times 10^{-06}$ | 0.00088 |

### 3.1. The Relative Importance of Input Variables

This paper developed an artificial neural network model to predict the cake moisture of pressure filtration. Figure 3 represents the relative importance of different input variables. Filtration time had the highest impact on cake moisture for both S1 and S2 datasets. The relative importance of the input variable was determined by the absolute value of the connection weight in the algorithm. The sign of the connection weight indicated the direction of the effect of the input variable on the output variable. A positive sign meant that the input variable had a direct effect on the output variable, while a negative sign meant that the input variable had an inverse effect on the output variable. However, it must be considered that the activation function and the bias term of the algorithm, as may affect the output as well.

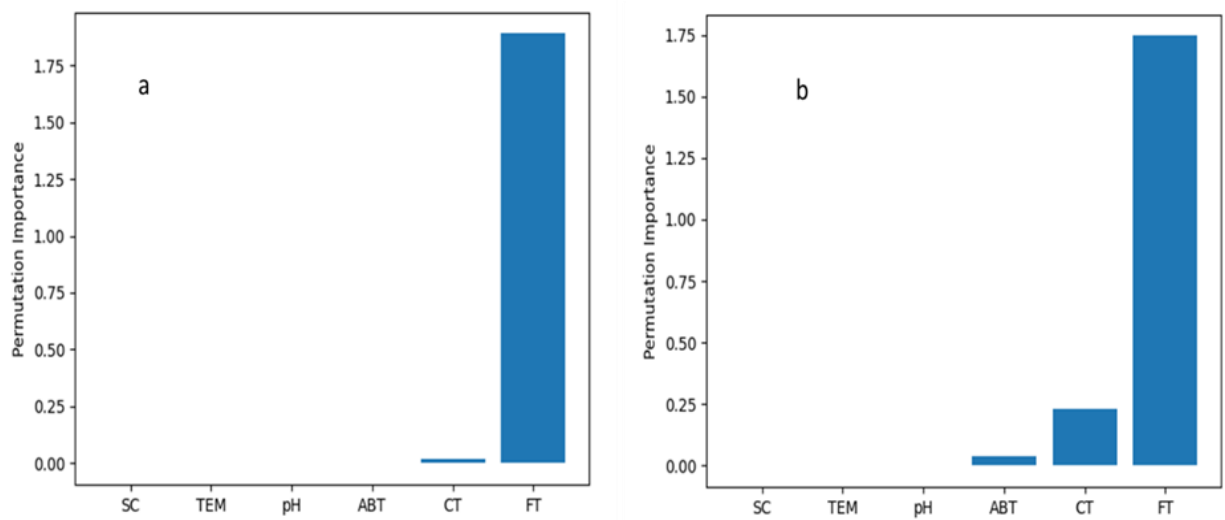

Figure 3 Relative importance of input variables from the ANN model: (a) S1, (b) S2

### 4. Conclusion

This study developed and applied an Artificial Neural Network (ANN) model to predict the cake moisture of pressure filtration in zinc production. The cake moisture was an important factor in the efficiency and profitability of the zinc leaching process. The study investigated the effects of seven operational parameters on cake moisture and conducted 288 tests using two types of fabrics, polypropylene, and polyester. The ANN model was trained and tested on the data and evaluated by $R^2$ (0.88 and 0.83), Mean Square Error (6.243×10$^{-07}$ and 1.086×10$^{-06}$), and Mean Absolute Error (0.00056 and 0.00088) metrics. The results showed that the ANN model had high accuracy and low error rates for both types of fabrics. The results also revealed that filtration time was the most significant parameter for cake moisture. The study demonstrated that the ANN model was a powerful and reliable tool for modeling and predicting the cake moisture of pressure filtration in zinc production. The study suggested that the ANN model could be used to optimize the processing conditions and increase the profits in zinc hydrometallurgy plants.

## Acknowledgments

The authors are grateful to the research and technology deputy of the University of Zanjan for the possibility of using the hydrometallurgical pilot plant.

## References

1. Haakana, T., Saxén, B., Lehtinen, L., Takala, H., Lahtinen, M., Svens, K., ... & Gongming, X. (2008). Outotec direct leaching application in China. *Journal of the Southern African Institute of Mining and Metallurgy*, *108*(5), 245-251.
2. Wang, J., Wang, Z., Zhang, Z., & Zhang, G. (2019). Zinc removal from basic oxygen steelmaking filter cake by leaching with organic acids. *Metallurgical and Materials Transactions B*, *50*, 480-490.
3. Wu, M., She, J. H., & Nakano, M. (2001). An expert control system using neural networks for the electrolytic process in zinc hydrometallurgy. *Engineering Applications of Artificial Intelligence*, *14*(5), 589-598.
4. Abdollahi, H., Noaparast, M., Shafaei, S. Z., Akcil, A., Panda, S., Kashi, M. H., & Karimi, P. (2019). Prediction and optimization studies for bioleaching of molybdenite concentrate using artificial neural networks and genetic algorithm. *Minerals Engineering*, *130*, 24-35.
5. Liu, X., Tian, S., Tao, F., & Yu, W. (2021). A review of artificial neural networks in the constitutive modeling of composite materials. *Composites Part B: Engineering*, *224*, 109152.
6. Ebrahimzade, H., Khayati, G. R., & Schaffie, M. (2018). Leaching kinetics of valuable metals from waste Li-ion batteries using neural network approach. *Journal of Material Cycles and Waste Management*, *20*(4), 2117-2129.
7. Li, Q. C., Li, D. X., & Chen, Q. Y. (2011). Prediction of pre-oxidation efficiency of refractory gold concentrate by ozone in ferric sulfate solution using artificial neural networks. *Transactions of Nonferrous Metals Society of China*, *21*(2), 413-422.
8. Kazemi, M., Moradkhani, D., & Alipour, A. A. (2023). Application of Random Forest and Support Vector Machine for Investigation of Pressure Filtration Performance, a Zinc Plant Filter Cake Modeling. *arXiv preprint arXiv:2307.14199*.
9. Werbos, P. (1974). Beyond regression: New tools for prediction and analysis in the behavioral sciences. *PhD thesis, Committee on Applied Mathematics, Harvard University, Cambridge, MA*.
10. Lek, S., & Park, Y. (2008). Artificial neural networks in Jorgensen SE, Fath B editors. Encyclopedia of Ecology.
11. Gardner, M. W., & Dorling, S. R. (1998). Artificial neural networks (the multilayer perceptron)-a review of applications in the atmospheric sciences. Atmospheric environment, 32(14-15), 2627-2636.